\documentclass[10pt, a4paper]{article}
\usepackage{lrec}
\usepackage{graphicx}
\usepackage{tabularx}
\usepackage{soul}

\usepackage{epstopdf}
\usepackage[latin1]{inputenc}

\usepackage{hyperref}
\usepackage{xstring}
\usepackage{amsmath}

\usepackage{csquotes}
\newcommand{\dq}[1]{\enquote{#1}}

\title{Two Huge Title and Keyword Generation Corpora of Research Articles}

\name{Erion \c{C}ano, Ond\v{r}ej Bojar}

\address{Institute of Formal and Applied Linguistics, Charles University \\
         Prague, Czech Republic \\
         \{cano, bojar\}@ufal.mff.cuni.cz\\}

\abstract{
Recent developments in sequence-to-sequence learning with neural networks have considerably improved the quality of automatically generated text summaries and document keywords, stipulating the need for even bigger training corpora. Metadata of research articles are usually easy to find online and can be used to perform research on various tasks. In this paper, we introduce two huge datasets for text summarization (OAGSX) and keyword generation (OAGKX) research, containing 34 million and 23 million records, respectively. The data were retrieved from the Open Academic Graph which is a network of research profiles and publications. We carefully processed each record and also tried several extractive and abstractive methods of both tasks to create performance baselines for other researchers. We further illustrate the performance of those methods previewing their outputs. In the near future, we would like to apply topic modeling on the two sets to derive subsets of research articles from more specific disciplines. \\ \newline  }

\begin{document}

\maketitleabstract

\section{Introduction}
\label{sec:intro}
The ongoing tendency towards data-driven solutions for more and more tasks such as MT (Machine Translation), TS (Text Summarization), KG (Keyword Generation), and other tasks related to natural languages has created incentives for crawling the Web to produce large text corpora of various types. Furthermore, recent open data initiatives of governments\footnote{\url{https://www.data.gov/open-gov}} and other institutions that encourage the publication of more data on the Web have induced the same effect. 
From academia, there are initiatives such as ArnetMiner \cite{tang2008arnetminer} that try to integrate existing scientific data from various resources in common networks for easier retrieval and exploitation. Among the various types of texts published in the Web, the metadata of research articles (e.g., titles, abstracts, keywords, etc.) are probably the easiest to find in large quantities, since they are usually not restricted. In fact, small corpora of research articles were used since the 90s to explore extractive KG \cite{Witten:1999:KPA:313238.313437,cogprints1802} and TS \cite{Mani:1997:MSG:1867406.1867503,Goldstein:2000:MSS:1117575.1117580} techniques. 
Research on these tasks has switched from the extractive paradigm to the recent abstractive one that is based on sequence-to-sequence learning with neural networks. The respective models are usually data-hungry, emphasizing the need for larger corpora in both TS and KG tasks.
In this paper, we first review the popular existing datasets used for TS and KG research. We later describe the processing steps we followed, starting from the retrieval of ArnetMiner OAG (Open Academic Graph) data collection to the creation of two novel and huge corpora: OAGSX\footnote{\url{http://hdl.handle.net/11234/1-3079}} and OAGKX.\footnote{\url{http://hdl.handle.net/11234/1-3062}} The first one contains more than 34 million records consisting of paper abstracts and titles. It is suitable for TS experiments (more specifically for title generation which is a form of TS). The second one contains roughly 23 million abstracts, titles, and lists of keywords and is best suited for KG experiments. 
The data samples in the two corpora were carefully examined and various statistics about the text lengths and the lexical similarities between abstracts, titles, and keywords are presented.   
We also explored the performance scores (ROUGE for TS and F$_1$@k for KG) of existing extractive and abstractive TS and KG solutions, trying them on evaluation minisets derived from OAGSX and OAGKX. According to our results, the recent abstractive methods based on sequence-to-sequence learning take a considerable time to train but perform better than the extractive methods on both tasks.  
To the best of our knowledge, our two data collections are the biggest of their kind that can be found online for free. We release them under the Creative Commons By 4.0 License.
As future work, we would like to perform topic recognition on the articles of the two collections. This may lead to the creation of many subsets of research articles data from more specific scientific disciplines. 

\section{Background}
\label{sec:background}

\subsection{Text Summarization}
\label{ssec:textsum}
Automatic TS research explores intelligent methods to compress text documents into shorter summaries that express the main ideas of the source.  
It is mostly driven by our need to have shorter and easy-to-read summaries of long documents for saving reading time. Sometimes, we also need to have summaries of conversation threads (e.g., emails or chat messages). Multi-document TS is important when we want concise information from a set of documents and summaries of conclusions from meetings (minuting) or other event discussions. Another type of summarization aims to create short client reviews about different aspects of certain products or services. Title generation is yet another form of TS which is about paraphrasing the content of a text to produce an appropriate title for it.  

There are two fundamental approaches for performing TS. The extractive way tries to select the most important and relevant parts from the source document and combines them to produce a shorter summary which is concise, coherent and readable. In this case, the target or output text contains verbatim copies of words or phrases taken from the source or input. The abstractive approach, on the other hand, learns to paraphrase the information required for the summary, instead of directly copying it from the source. This is somehow better, but the methodology is more complex and requires more resources.
The TS research of the 90s and early 00s was mostly based on extractive methods. The respective techniques used unsupervised learning \cite{Goldstein:2000:MSS:1117575.1117580,Barzilay97usinglexical}, supervised learning \cite{Wong:2008:ESU:1599081.1599205,DBLP:conf/ntcir/Fukumoto04} or graph methods \cite{Erkan_2004,Mani:1997:MSG:1867406.1867503} to select the most important lexical units from the source documents. 

The abstractive approach has become popular in recent years, following the progress in sequence-to-sequence learning with neural networks (the encoder-decoder framework). LSTM neural networks \cite{Hochreiter:1997:LSM:1246443.1246450} are combined and enhanced with advanced mechanisms like the attention of \newcite{DBLP:journals/corr/BahdanauCB14} for a more effective learning of the alignments between the text sequences. Attention allows the model to focus on different segments of the input during generation and was successfully used by \newcite{D15-1044} to summarize news articles. The problem of unknown words (not seen in source texts) was also mitigated by the copying technique \cite{gu-EtAl:2016:P16-1,gulcehre-etal-2016-pointing}. Furthermore, the coverage \cite{P16-1008} and intra-attention \cite{DBLP:journals/corr/PaulusXS17} mechanisms were proposed to alleviate word repetitions in the summaries, a notorious problem of the encoder-decoder models. 

Scoring results were pushed even further just recently by mixing reinforcement learning concepts such as policy gradient \cite{Rennie2017SelfCriticalST} into the encoder-decoder architecture. It optimizes the learning objective (higher summarization score) and still keeps an appropriate quality of the produced summaries. A recent performance comparison of various abstractive TS methods can be found in \newcite{cano-bojar-2019-efficiency}.

\subsection{Keyphrase Generation}
\label{ssec:keygen}
Keyphrase generation is the process of analyzing a document and producing sets of one or a few words (keywords or keyphrases, used interchangeably) that best represent its main concepts or topics. These keywords are frequently utilized nowadays to annotate digital objects (e.g., research articles, books, product descriptions, etc.) and quickly find them in digital libraries, online stores, etc. A keyword string is a concatenation of several keywords (commas or semicolons are typically used as separators) attached to one of those objects. 
The need to process large amounts of documents with missing keywords created incentives for research in automatic KG since the 90s. The popular supervised learning algorithms of that time were used by authors like \newcite{Turney:2000:LAK:593957.593993} or \newcite{Witten:1999:KPA:313238.313437} in combination with lexical features to extract keywords from the documents. Furthermore, graph-based methods \cite{Rose2010,Wan:2008:SDK:1620163.1620205} or other unsupervised KG methods \cite{10.1007/978-3-319-76941-7_80,Nart2014ADI} were proposed later in the 00s.
The above extractive KG solutions were very successful because of their simplicity and execution speed. However, extractive KG suffers from a serious inherent handicap: its inability to produce absent keywords (keywords not appearing in the source text). \newcite{P17-1054} analyzed the author's keywords in popular corpora. They observed that absent and present (keywords that also appear in the source text) keywords assigned by paper authors are almost equally frequent. It is thus a serious drawback to completely ignore the absent keywords.
The recent advances in language representation \cite{2013arXiv1301.3781M,pennington2014glove} and sequence-to-sequence learning \cite{DBLP:journals/corr/BahdanauCB14,NIPS2017_7181} motivated several researchers like \newcite{P17-1054} or \newcite{8438457} to explore abstractive KG in the context of the encoder-decoder framework. The encoder-decoder network structures were initially utilized to perform MT and got quick adoption on similar tasks like TS and KG that are also based on the sequence-to-sequence transformation between source and target texts. Furthermore, same as in TS research, various reinforcement learning concepts like adaptive rewards that are being explored are raising the performance scores even higher \cite{chan-etal-2019-neural}. Abstractive KG is now a vibrant research direction with more than a dozen of publications only in the last three years. More comprehensive surveys of KG literature can found in other recent publications like \cite{doi:10.1002/widm.1339} and \cite{cano-bojar-2019-fruct}.

\subsection{Scientific Article Data Sources}
\label{ssec:sciart}
The current hype of deep neural networks has created strong incentives for producing data collections by crawling the web. The richest sets of language resources are used for machine translation \cite{resnik-smith-2003-web,TIEDEMANN12.463,mahata-etal-2016-wmt2016,shi2005a} and for sentiment analysis  \cite{6468031,nlpinai19,maas-EtAl:2011:ACL-HLT2011,doi:10.1108/DTA-03-2018-0017,zaframulti}. They are mostly driven by the information technology giants that continuously improve their language-related applications and marketing companies to understand customers' perceptions about various online products. 

TS and KG research of the 90s and early 00s was mostly based on extractive methods that did not rely on big training corpora. Things gradually changed in the late 00s with the rising popularity of the encoder-decoder framework. The current TS and KG methods are also highly dependent on the big language corpora since they are mainly based on sequence-to-sequence learning with neural networks. Some of the most popular corpora in TS and KG literature are presented in Table~\ref{tab:datasets}. 
One of the first big datasets was the annotated English Gigaword \cite{Napoles:2012:AG:2391200.2391218} used for abstractive TS by \newcite{D15-1044}. It contains about nine million news articles and headline summaries. Each headline was paired with the first sentence of the corresponding article to create the training base for the experiments.  
Newsroom \cite{N18-1065} is a very recent and heterogeneous bundle of about 1.3 million news articles. It contains writings published from 1998 to 2017 by 38 major newsrooms. 
Another recent dataset of news articles is CNN/Dailymail of \newcite{K16-1028}. It has become the most popular corpus for text summarization experiments. This dataset provides a rich collection of news articles and the corresponding multi-sentence summaries (news highlights). It is thus very suitable for training and testing summarization models of longer texts.
DUC-2004 is another dataset that was originally created for the Document Understanding Conference.\footnote{\url{https://duc.nist.gov/duc2004/}} It has been mostly used as an evaluation baseline, given its small size. It consists of 500 document-summary pairs curated by human experts.

Besides using news articles, it is also possible to exploit texts of scientific articles for TS research.
In fact, those kinds of texts have been used since long ago to conduct KG research. There are many relatively small datasets of scientific publications and the corresponding keywords that have been used for many years to test extractive or graph-based KG methods.   
One of the most popular KG datasets is Inspec released by \newcite{hulth2003improved}. It consists of 2000 paper titles (1500 for training and 500 for testing), abstracts and keywords from journals of Information Technology, published from 1998 to 2002.
\begin{table}[t]
\begin{center}
\begin{tabular}{| l | c | c | c |}
\hline
\bf Reference & \bf Name & \bf Content & \bf \# Docs \\
\hline
\newcite{Napoles:2012:AG:2391200.2391218} & Gigaword & News & 9\,M \\
\newcite{N18-1065} & Newsroom & News & 1.3\,M \\
\newcite{K16-1028} & CNN/DM & News & 287\,K \\
\href{https://duc.nist.gov/duc2004/}{Hyperlink} & DUC-2004 & News & 500 \\
\newcite{hulth2003improved} & Inspec & Papers & 2000 \\
\newcite{krap2010} & Krapivin & Papers & 2304 \\
\newcite{S10-1004} & SemEval & Papers & 244 \\
\newcite{P17-1054} & KP20k & Papers & 567\,K \\
\newcite{DBLP:journals/corr/abs-1804-08875} & tit-gen & Papers & 900\,K \\
\newcite{DBLP:journals/corr/abs-1804-08875} & abs-gen & Papers & 5\,M \\
\hline
\end{tabular}
\caption{\label{tab:datasets}Summary and keyword generation datasets}
\end{center}
\end{table}

\newcite{krap2010} released another collection of papers that has been frequently used in the literature. It consists of 2304 computer science papers published by ACM from 2003 to 2005. The advantage of this dataset is the availability of the full paper texts together with the corresponding metadata.
A smaller dataset is SemEval of \newcite{S10-1004} that was originally created for the Semantic Evaluation task. It contains 244 papers that belong to conference and workshop proceedings.  

A few years ago, \newcite{P17-1054} released KP20k which is today the most popular KG dataset. It contains 567830 Computer Science articles, 527830 used for training, 20\,K for validation and 20\,K for testing. This dataset has been used for training and comparing various recent abstractive KG methods.
\newcite{DBLP:journals/corr/abs-1804-08875} raised the data sizes even more by retrieving many scientific papers from libraries of biomedical research.\footnote{\url{https://www.nlm.nih.gov}} The authors derived and released two big (900\,K and 5\,M) corpora for TS (predicting abstracts from paper bodies) and title generation (predicting titles from abstracts).

Crawling public digital libraries or websites for text resources is an ongoing trend. ArnetMiner \cite{tang2008arnetminer} is an initiative to integrate scientific data (publications, researcher profiles and more) from various resources in a common and unified network. A derivative product is the OAG data collection of scientific publications \cite{Sinha:2015:OMA:2740908.2742839}. Each record is a JSON line with publication metadata like \emph{authors}, \emph{title}, \emph{abstract}, \emph{keywords}, \emph{year} and more. In the following section, we describe the processing steps we performed on OAG collection to derive OAGSX and OAGKX datasets.    

\section{OAGSX and OAGKX Corpora}
\label{sec:oagksx}
For producing large TS and KG text collections, we utilized the text fields of the OAG bundle. From that same article set, we filtered the records containing at least the \emph{title} and the \emph{abstract} for OAGSX and those with the \emph{title}, \emph{abstract}, and \emph{keywords} for OAGKX. We dropped the duplicate entries in each of our two collections. As a result, the samples inside each of the corpora are unique (there is still overlapping between OAGSX and OAGKX samples, since they were both derived from the OAG collection). An automatic language identifier\footnote{\url{https://pypi.org/project/langid}} was used to remove the records with abstracts not in English. We also cleared the messy symbols and lowercased everything. Finally, Stanford CoreNLP \cite{manning-EtAl:2014:P14-5} was used to tokenize the \emph{title} and \emph{abstract} texts.

\begin{table}[t]
\begin{center} 
\begin{tabular}{|l|l|}
\hline
\bf Attribute & \bf \quad Title \qquad~ Abstract \\
\hline
Total & ~~\, 449\,M \qquad\quad 6\,B \\
\hline
Min\,/\,Max & \quad 3\,/\,25 \qquad~\, 50\,/\,400 \\
\hline
Mean\,(Std) & 13.1\,(5.1) ~~~ 182.2\,(89.2) \\
\hline
Jindex & \qquad 6.7\,\%\,(3.9\,\%) \\
\hline
Overlap & \qquad 77\,\%\,(18\,\%) \\
\hline
Total size & 34\,408\,509 title-abstracts \\
\hline
\end{tabular}
\caption{\label{tab:sstat}Token statistics of OAGSX}
\end{center}
\end{table}
\begin{table}[t]
\begin{center}
\begin{tabular}{|l|l|}
       \hline
       \bf Attribute & \bf ~~~~Title \qquad Abstract ~~~~~ Keywords \\
       \hline
       Total & ~~ 290\,M \qquad~~ 4\,B \qquad\quad~~ 270\,M \\
       \hline
       Min\,/\,Max & \quad 3\,/\,25 \qquad\, 50\,/\,400 \qquad~ 2\,/\,60 \\
       \hline
       Mean\,(Std) & 12.8\,(4.9) ~ 175.1\,(86.5) ~~~ 11.9\,(7.5) \\
		\hline
       Jindex & \quad~ 7.1\,\%\,(4\,\%) \quad~ 6\,\%\,(4.8\,\%) \\
		\hline
       Overlap & ~~~\, 78\,\%\,(17\,\%) \quad~ 68\,\%\,(25\,\%) \\
		\hline
       Total size & 22\,674\,436 title-abstract-keywords \\ 
		\hline
\end{tabular}
\caption{\label{tab:kstat}Token statistics of OAGKX}
\end{center}
\end{table}
\begin{table}[t]
\begin{center}
\begin{tabular}{|l|c|}
       \hline
      \bf Attribute & \bf Value \\
       \hline
       Total & 133\,295\,056 \\ 
       \hline
       Min\,/\,Max & 2\,/\,12 \\
		\hline
       Mean\,(Std) & 5.9\,(3.1) \\
		\hline
       Present & 52.7\,\%\,(28.3\,\%) \\
		\hline
       Absent & 47.3\,\%\,(28.3\,\%) \\
       \hline
\end{tabular}
\caption{\label{tab:kstat2}Keyword statistics of OAGKX}
\end{center}
\end{table}
After the preprocessing steps, we observed the size and token lengths of the records. Since there were many outliers (e.g., records with very long or very short abstracts), we removed all records with a title not in the range of 3-25 tokens and abstract not within 50-400 tokens. In the case of OAGKX, we also removed samples with keyword string not in the range of 2-60 tokens or 2-12 keywords. After this, OAGSX was reduced to a total of about 34.4 million records. OAGKX, on the other hand, shrank to about 22.6 million records. 
Some further statistics of the two final datasets are presented in Tables~\ref{tab:sstat} and \ref{tab:kstat}. In the case of OAGSX, the average title and abstract lengths are about 13.1 and 182.2 tokens respectively (standard deviation is always given in parenthesis). The corresponding values in OAGKX are 12.8 and 175.1 (slightly lower). For OAGKX we also see that the keyphrase strings contain 11.9 tokens on average. 
We also wanted to observe the lexical similarity between the titles and abstracts. One way for this is to compute the Jaccard similarity (Jindex in Tables~\ref{tab:sstat} and \ref{tab:kstat}) of the whole token sets using the following equation:
\begin{equation}
\label{eq:jacardindex}
J(A, B) = \frac{|T \cap A|}{|T \cup A|} = \frac{|T \cap A|}{|T| + |A| - |T \cap A|}
\end{equation}
where $T$ is the set of unique tokens in the title and $A$ is the set of unique tokens in the abstract. In OAGSX, the Jaccard similarity between abstracts and titles is 6.7\,\%. In OAGKX, it is 7.1\,\% between the abstracts and titles and 6\,\% between abstracts and keyword strings. Another indicator is the overlap $o(s, t) = \frac{|\{s\} \cap \{t\}|}{|\{t\}|}$ which represents the fraction of unique target tokens $t$ (e.g., in the \emph{title} or in the \emph{keyword string} excluding punctuation symbols) that overlap with a source token (e.g., in \emph{abstract}) $s$. The overlaps between titles and abstracts are very similar (77\,\% and 78\,\%) in both datasets. In the case of OAGKX, the overlap between abstracts and keyword strings is 68\,\%.
We further analyzed the keyword distribution in OAGKX (Table~\ref{tab:kstat2}). There is a total of about 133 million keywords, with an average of 5.9 keywords per article. In abstractive KG experiments, it is also important to know the distribution of present and absent keywords. The present rate $p(s, k) = \frac{|k~\cap~s|}{|k|}$ is the fraction of the keywords $k$ that also appear in the source text $s$. This is similar to the overlap, with the difference that there might be token repetitions within each counted keyword. The absent rate $a(s, k) = \frac{|k| - |k~\cap~s|}{|k|}$ is its complement or the fraction of keywords $k$ that do not appear in the source text $s$. From Table~\ref{tab:kstat2} we see that the present and absent keywords in OAGKX are almost evenly distributed (52.7\,\% and 47.3\,\% each). This observation is in line with that of \newcite{P17-1054}, emphasizing once again the importance of the absent keywords.   

Another interesting exploration we wanted to perform was the identification of the topics (or research domains) in each dataset record to report the corresponding statistics. This could lead to the creation of many subsets of OAGSX and OAGKX with scientific articles from more specific disciplines (clustering together the articles from the same research direction). Unfortunately, topic modeling was not easy to perform on OAGSX and OAGKX, given the huge size of the two corpora and our limited computational resources. It thus remains a potential future work. 

We still inspected a few of the samples from each dataset manually. Their texts mostly belong to papers from biomedical disciplines but there are also papers about psychology, geology, or various technical directions. To our best knowledge, OAGSX and OAGKX are the largest available collections of scientific paper metadata that can be used for TS and KG experiments. Their importance is thus twofold: (i) They can supplement existing collections if more training samples are required. (ii) They can serve as sources for deriving article subsets of more specific scientific disciplines or domains.       

\section{Evaluation Experiments}
\label{sec:eval}
We tried various extractive and abstractive methods for TS and KG on evaluation subsets from two corpora. In the following sections, we report the achieved performance scores of the automatic evaluation process. We also illustrate the output of each method with examples.   

\subsection{Title Generation}
\label{ssec:evaltitgen}
For the title generation experiments, we formed three evaluation subsets from OAGSX: a training set of 1 million samples, a validation set of 10 thousand samples and a test set of 10 thousand samples. To reduce the vocabulary size (important for abstractive text summarizers), we further replaced number patterns with the \# symbol in each of them.  
The most simple and raw baseline we used is Random-k (Random-1 in our case) which splits the source text into sentences and randomly picks $k$ of them as its summary. In our case, since we are generating the title of the articles, we randomly pick only one of the abstract sentences as the predicted title. Random-1 can be considered as the lowest scoring boundary since it uses no intelligence at all. 
Another popular baseline is Lead-k (Lead-1 in our case). It is based on the concept of \dq{summary lead}, which concisely explains the main idea of a text in its first sentence or first few sentences. Lead-1 picks the first sentence from the source text to generate its title. 
LexRank is a stochastic graph-based method for assessing the importance of textual units in a source text \cite{Erkan_2004}. When used to perform extractive TS, it computes the importance of those units using the concept of eigenvector centrality in the graph. The top $k$ units (the top sentence in this case) are returned as the best summary of the document.  

One of the abstractive text summarizers we used is PointCov of \newcite{P17-1099} which is based on the encoder-decoder framework. In each decoding step, it implements the pointing/copying mechanism \cite{gu-EtAl:2016:P16-1,gulcehre-etal-2016-pointing} to compute a generation probability. The latter is used to decide whether the next word should be predicted or directly copied from the source sequence. Another feature is the implementation of the coverage mechanism \cite{P16-1008} which helps to avoid word repetitions in the target sequence. We trained PointCov with a hidden layer of 256 dimensions and word embeddings of 128 dimensions.

The other abstractive summarizer we picked is the Transformer model that represents one of the most important achievements in sequence-to-sequence learning of the last years \cite{NIPS2017_7181}. It is totally based on the attention mechanism, removing all recurrent or convolutional structures. Although it was primarily designed for MT, the Transformer can also work for text summarization. It basically learns the alignments between the input (source) texts and the output (target) summaries. As documented by \newcite{cano-bojar-2019-efficiency}, Transformer reveals the highest data efficiency scores on the popular TS datasets. We used the Transformer model with four layers in both encoder and decoder blocks, 512 dimensions in each layer, including the embedding layers, 200\,K training steps, and 8000 warm-up steps. Both PointCov and Transformer were trained with Adam optimizer \cite{Adam} using $\alpha = 0.001,~\beta_1 = 0.9,~\beta_2 = 0.999$ and $\epsilon = 10^{-8}$ and mini-batches of 16 training samples. We used two NVIDIA GTX 1080Ti GPUs at once for the training process. 

Random-1, Lead-1, and LexRank (the three extractive methods) were directly applied in the test set of 10 thousand examples.
For PointCov and Transformer, we used all the three evaluation subsets. ROUGE-1, ROUGE-2, and ROUGE-L scores \cite{Lin:2004} were computed by comparing the title outputs of each method with the titles of the original papers. The results are presented in Table~\ref{tab:scoresOAGSX}.   
As we expected, Random-1 is the worst in all three ROUGE scores. Lead-1 performs well, reaching a peak score of 33.8\,\% in ROUGE-1. It is actually slightly better than LexRank in all the three metrics. Transformer and PointCov, which are the two abstractive neural networks we tried, perform better than the three extractive methods. They achieve similar results, but the Transformer leads with a peak score of 37.27\,\% in ROUGE-1. It is also important to note that the three extractive methods took only a few minutes to produce the outputs. PointCov and Transformer, on the other hand, required 3 -- 4 days for the training.  

An abstract, its author's title, and the titles predicted by the above five methods are illustrated in Table~\ref{tab:qualOAGSX}. As we can see, all the methods have generated titles that are longer than those of the authors. The titles of Lead-1 and LexRank are very similar, both based on the first sentence of the abstract. The transformer has produced a very long title with an unfinished sentence. This problem could be fixed by using a lower value for the length of the target text. PointCov has generated a shorter sentence than the Transformer, but it is not very coherent.  
\begin{table}[ht]
\begin{center}
\begin{tabular}{| l | c c c |}
\hline
\bf Method & $\boldsymbol{R_1}$ & $\boldsymbol{R_2}$ & $\boldsymbol{R_L}$ \\
\hline
Random-1 & 22.67 & 8.02 & 18.44 \\
Lead-1 & 33.83 & 16.8 & 28.14 \\
LexRank & 29.4 & 12.83 & 24.03 \\
PointCov & 36.12 & 18.88 & 30.21 \\
Transformer & 37.27 & 19.12 & 30.78 \\
\hline
\end{tabular}
\caption{\label{tab:scoresOAGSX}Results on OAGSX}
\end{center}
\end{table}

\begin{table}[!t]
\begin{center}
\begin{tabular}{| p{77mm} |}
\hline
\textbf{Abstract:} the central bank 's lender of last resort role was developed by a series of authors in the very late eighteenth and through the nineteenth centuries . it was tested in practice in a number of countries and was found to be effective in providing monetary stability in the face of adverse shocks . there have recently been attempts to broaden the role to make the central bank responsible for the stability of asset markets , or for protecting individual banks and there have recently also been claims that an international lender of last resort is necessary . this article considers and rejects these proposed extensions to the classic lender of last resort role \\ [0.07ex]
\hline

\textbf{Author's title:} the lender of last resort reconsidered  \\ [0.07ex]
\hline

\textbf{Random-1 title:} this article considers and rejects these proposed extensions to the classic lender of last resort role \\ [0.07ex]
\hline

\textbf{Lead-1 title:} the central bank 's lender of last resort role was developed by a series of authors in the very late eighteenth and through the nineteenth centuries \\ [0.07ex]
\hline

\textbf{LexRank title:} the central bank 's lender of last resort role was developed in the late eighteenth and through the nineteenth centuries \\ [0.07ex]
\hline

\textbf{PointCov title:} the central bank 's lender and its implications for the stability of asset comparative analysis \\ [0.07ex]
\hline

\textbf{Transformer title:} the central bank 's lender of last resort role and its implications for the stability of asset markets : a comparative analysis of the central and in the lender of the \\ [0.07ex]
\hline
\end{tabular}
\caption{\label{tab:qualOAGSX}KE scores on OAGKX}
\end{center}
\end{table}

\subsection{Keyphrase Generation}
\label{ssec:evalkeygen}
We ran similar experiments on three evaluation sets derived from OAGKX: a training set of 631705 samples, a validation set of 10 thousand samples and a test set of 10 thousand samples. Once again, we tried and compared both extractive and abstractive KG methods.
We used TopicRank of \newcite{I13-1062} which is a popular graph-based extractive method that makes use of the PageRank algorithm \cite{Brin:1998:ALH:297810.297827}. It first uses clustering to group lexical units of the same topic. Then, it uses the graph-based ranking algorithm to score each topic cluster that is formed. At the end, one keyword is picked from each of the ranked clusters.  

\begin{table}[!t]
\begin{center}
\begin{tabular}{|l| l l c|} 
\hline
\bf Method & \bf F\textsubscript{1}@5 & \bf F\textsubscript{1}@7 & \bf F\textsubscript{1}@10 \\ [0.1ex] 
\hline
TopicRank & 17.12 & 20.81 & 20.75 \\ [0.07ex]
RAKE & 16.36 & 18.84 & 18.91 \\ [0.07ex]
Maui & 24.58 & 23.49 & 23.6 \\ [0.07ex]
CopyRNN & 28.15 & 28.93 & 28.96 \\ [0.07ex]
CovRNN & 27.76 & 29.15 & 29.04 \\ [0.07ex]
\hline
\end{tabular}
\caption{\label{tab:scoresOAGKX}KE scores on OAGKX}
\end{center}
\end{table}
\begin{table}[!t]
\begin{center}
\begin{tabular}{| p{77mm} |}
\hline
\textbf{Abstract:} a complex polysaccharide accumulation was observed in the central nervous system ( cns ) of rats treated with d-penicillamine similar to lafora-like bodies . they have histochemical similarities comparable to bodies described in previous studies of lafora disease . the clinical usefulness of d-penicillamine has been limited by many side effects including renal damage . it is suggested that , in addition to d-penicillamine nephropathy , there are toxic effects of this drug on the cns \\ [0.07ex]
\hline

\textbf{Title:} polysaccharide accumulation in the central nervous system of d-penicillamine treated rats \\ [0.07ex]
\hline

\textbf{Author's keywords:} polysaccharide , central nervous system , side effect , d-penicillamine , lafora-like bodies , nephropathy \\ [0.07ex]
\hline

\textbf{TopicRank keywords:} central nervous system , d-penicillamine , accumulation , polysaccharide accumulation , cns , d-penicillamine effect , drug , lafora bodies , clinical , rats \\ [0.07ex]
\hline

\textbf{RAKE keywords:} clinical , polysaccharide , central nervous , rats , polysaccharide accumulation , lafora disease , renal damage , accumulation , lafora, cns  \\ [0.07ex]
\hline

\textbf{Maui keywords:} central , central system , system , d-penicillamine , polysaccharide accumulation , polysaccharide , accumulation , lafora-like , lafora, bodies  \\ [0.07ex]
\hline

\textbf{CopyRNN keywords:} central nervous , d-penicillamine , side effect , side , newborn rats , rat ileostomy , pregnant rats,  nephropathy , mortality of rats , mortality  \\ [0.07ex]
\hline

\textbf{CovRNN keywords:} side effect , central nervous , polysaccharide , rats , lafora bodies , polysaccharide effect , d-penicillamine , polysaccharide-derived , albino rats , trinitrobenzene sulfonic acid \\ [0.07ex]
\hline
\end{tabular}
\caption{\label{tab:qualOAGKX}KE scores on OAGKX}
\end{center}
\end{table}
RAKE proposed by \newcite{Rose2010} is one of the fastest available methods for extractive KG. It first removes punctuation symbols together with the stop words of the specified language and then creates a graph of word co-occurrences. Candidate words or phrases are scored based on the degree and frequency of each word vertex in the graph. The $k$ top-scoring candidates are returned as keywords.
We also used Maui \cite{medelyan2009human}, a supervised extractive method that uses lexical features and bagged decision trees to predict whether a candidate phrase is a keyword or not.   
CopyRNN \cite{P17-1054} was the first abstractive KG method based on the encoder-decoder framework. Authors implemented the copying mechanism to balance between extracting present phrases from the source text with the generation of absent phrases. This work was followed by several recent studies that improve KG with various additional mechanisms.   
Finally, the last method we tried is CovRNN \cite{8438457} which is very similar to CopyRNN. It tries to avoid the repeated keywords during generation by considering the correlation between the produced keywords at each generation step. This is achieved by implementing the coverage mechanism of \newcite{tu-etal-2016-modeling}.  

We applied TopicRank and RAKE on the test set of 10 thousand records. Because of its memory limitations, Maui was trained on the first 30 thousand samples from the training set and tested on the test set. For CopyRNN and CovRNN, we used the full sizes of the three evaluation sets. For the comparison, we used F$_1$ scores of the full matches between the author's keywords and the top $k$ keywords returned by each method. Given that each data sample has a variable-length keyword string, we picked the values 5, 7 and 10 for the $k$ parameter. The obtained results are shown in Table~\ref{tab:scoresOAGKX}. 

The first thing we can notice from the results is the fact that F$_1@7$ and F$_1@10$ scores are very similar to each other in each case. This is probably because few data samples contain more than 7 keywords in their keyword string (the average was 5.9). We also see that CopyRNN and CovRNN perform significantly better than the first three extractive methods. They achieve very similar scores in the three metrics. The peak score of 29.15\,\% is reached by CovRNN on F$_1@7$. From the three extractive methods, Maui performs better than the other two. TopicRank performs slightly better than RAKE. Once again, the training of the abstractive methods based on neural networks took about 3 days whereas the results of the extractive approaches (with the exception of Maui which was trained in few hours) were obtained in few minutes.    
The source texts and the produced keywords (top ten) of a data sample are shown in Table~\ref{tab:qualOAGKX}. Apparently, both extractive and abstractive predictions are grammatically correct. However, few of the generations represent full keyword matches. There is also a considerable number of partial matches. The first four methods have produced certain word repetitions. We can also observe some \dq{novel} (thou incorrect) phrases like \dq{mortality of rats} or \dq{trinitrobenzene sulfonic acid} that are produced by CopyRNN and CovRNN. 

\section{Conclusion}
\label{sec:conclusions}
Today, we can find uncountable research article data that are freely available in digital libraries. Many relatively small collections of those data are frequently used to run text summarization and keyword generation experiments. 
In this paper, we described the steps we followed to process Open Academic Graph data and prepare two huge corpora: OAGSX of more than 34 million abstracts and titles that can be used for text summarization and OAGKX of about 23 million abstracts, titles, and keyword strings that can be used for keyword generation. To our best knowledge, these corpora of scientific paper metadata are the biggest freely available online. 
We also performed several experiments applying extractive and abstractive TS and KG methods on their subsets to help establish performance benchmarks that could be valuable to other researchers. 
In the future, we plan to apply topic modeling on the two collections for deriving many subsets of research articles from more specific scientific disciplines. 

\section{Acknowledgements}
\label{sec:ack}
This research work was supported by the project No. CZ.02.2.69/0.0/0.0/16\_027/0008495 (International Mobility of Researchers at Charles University) of the Operational Programme Research, Development and Education, the project no. 19-26934X (NEUREM3) of the Czech Science Foundation and ELITR (H2020-ICT-2018-2-825460) of the EU. 

\section{Bibliographical References}
\label{main:ref}

\bibliographystyle{lrec}
\bibliography{lrec}


\end{document}